%% file: emnlp2022.tex
\pdfoutput=1

\documentclass[11pt]{article}

\usepackage{EMNLP2022}

\usepackage{times}
\usepackage{latexsym}

\usepackage[T1]{fontenc}

\usepackage[utf8]{inputenc}

\usepackage{graphicx}
\usepackage{multirow}
\usepackage{color}
\usepackage{subfigure}
\usepackage{color}
\usepackage{tcolorbox}
\usepackage{CJK}
\usepackage{adjustbox}

\usepackage{microtype}

\usepackage{inconsolata}

\title{Time Will Change Things: An Empirical Study on Dynamic Language Understanding in Social Media Classification}

\author{Yuji Zhang, Jing Li\\
  Department of Computing, \\
  The Hong Kong Polytechnic University, \\
  HKSAR, China \\
  \texttt{yu-ji.zhang@connect.polyu.hk}~~\texttt{jing-amelia.li@polyu.edu.hk}\\}

\begin{document}
\maketitle

\begin{abstract}
Language features are ever-evolving in the real-world social media environment.
Many trained models in natural language understanding (NLU), ineffective in semantic inference for unseen features, might consequently struggle with the deteriorating performance in dynamicity.
To address this challenge, we empirically study social media NLU in a \textbf{dynamic setup}, where models are trained on the past data and test on the future.
It better reflects the realistic practice compared to the commonly-adopted static setup of random data split.
To further analyze model adaption to the dynamicity, we explore the usefulness of leveraging some unlabeled data created after a model is trained. 
The performance of unsupervised domain adaption baselines based on auto-encoding and pseudo-labeling and a joint framework coupling them both are examined in the experiments.
Substantial results on four social media tasks
imply the universally negative effects of evolving environments over classification accuracy, while auto-encoding and pseudo-labeling collaboratively show the best robustness in dynamicity.

\end{abstract}

\input{sections/introduction.tex}
\input{sections/related-work.tex}

\input{sections/framework.tex}

\input{sections/datasets-setup.tex}
\input{sections/results-analysis.tex}

\input{sections/conclusion.tex}

\section*{Limitations}



In the following, we summarize the limitations of our study in three points.


First, this work is studied in a relatively idealistic setup where we assume the availability of trans-data. However, in the real world, there may be challenges that hinder us from acquiring the related data as the trans-data. 
Second, in real world, VAE has to learn word statistics from noisy data consisting of a broad range of topics. The noisy data may pose challenges for VAE to capture the word co-occurrence pattern. We should provide more contextual information and knowledge to make better use of the trans-data in the future. 
Third, according to our analysis on label effect, PL relies heavily on its pseudo labels accuracy. Future work may consider how to encode a robust learning with imperfect pseudo labeling.

\section*{Ethics Statement}
We declare this work will not present any ethical problem.
In our empirical study, datasets \textsc{Stance}, \textsc{Fake-News}, and \textsc{Hate-Speech} are publicly  available with well-reputed previous work, where the ethical concerns have been addressed by the authors of these original papers.

For the data collection of \textsc{Hashtag} and missing data in previous benchmarks, Twitter official API was employed strictly following the Twitter terms of use.
The newly gathered data was thoroughly examined to exclude any possible ethical risks, e.g.,
toxic language and user privacy. 
We also conducted data anonymization  in pre-processing by removing user identities and replacing @mention with a generic tag.


\bibliography{anthology,custom}
\bibliographystyle{acl_natbib}




\end{document}

%% file: sections/introduction.tex
\section{Introduction}

The advance of natural language understanding (NLU) 
automates the learning of text semantics, exhibiting the potential to broadly benefits social media applications.
As shown in the previous work~\cite{tong-etal-2021-recent, DBLP:conf/uemcom/HeidariJ20, DBLP:journals/hcis/SalminenHCJAJ20}, the pre-trained  
models from the BERT family~\cite{devlin-etal-2019-bert, DBLP:journals/corr/abs-1907-11692, nguyen-etal-2020-bertweet} have championed the benchmark results in many social media tasks.
Nevertheless, 
will good benchmark results also
indicate good real-world performance on social media?

In view of our dynamic world, 
it is not hard to envision an ever-evolving environment on social media, which is shaped by what and how things are discussed there in real time. 
As a result, language features, formed by word patterns appearing 
there, might also rapidly change over time.
However, many trendy NLU models, including the state-of-the-art (SOTA) ones based on pre-training, demonstrate compromised empirical results facing shifted features~\cite{hendrycks-etal-2020-pretrained}.
The possible reason lies in the widely-argued limitation of existing NLU solutions on inferring meanings of new or shifted features compared to what the models have seen in the training data~\cite{DBLP:journals/corr/abs-1810-08750, DBLP:journals/corr/abs-1907-02893, DBLP:conf/icml/CreagerJZ21, DBLP:conf/aaai/Shen0ZK20, DBLP:conf/icml/LiuH00S21}.

Consequently, the dynamic social media environment in the realistic scenarios will continuously challenge a trained NLU model with timely-increasing unseen features~\cite{nguyen2012predicting}, further resulting in a deteriorating performance as time goes by.
To better illustrate this challenge, we take the task and dataset of Twitter stance detection for COVID-19 topics as an example 
\cite{glandt-etal-2021-stance}.
Two models based on LSTM and BERT are trained on the past data and test on five datasets with varying spans to the training set. 
The setup and results are detailed in Figure \ref{fig:intro-case} (right).

\input{figures/intro-case}

Both models exhibit dropping accuracy scores over time, implying the concrete challenge for them to tackle dynamicity. 
To further analyze the reasons, we employ 
variational auto-encoder (VAE) ~\cite{DBLP:journals/corr/KingmaW13} to learn the latent topics (word clusters) from varying test sets
and display the words indicating the largest correlation with each cluster in Figure \ref{fig:intro-case} (left).
It is observed that users' discussion points change over time,
where the focus 
gradually shifted from the concern to the virus
itself (indicated by words like ``\textit{Mask}'', ``\textit{Immune Compromise}'', ``\textit{Lock Down}'') to the disappointment to the former US President Trump (e.g., ``\textit{Trump Land Slid}'' and ``\textit{Lying Trump}'').
Because of the topic evolution, it might not be easy for models trained with the $t_0$ data to connect the later-gathered ``\textit{Trump}''-patterns to an ``against'' stance for topics related to his COVID-19 policies.



To empirically examine 
how dynamicity affects NLU performance, we experiment in a \textbf{dynamic setup}: the data is split with an absolute time, where the messages posted beforehand are used for training while those afterwards are for test.
On the contrary, most social media benchmarks adopt the \textbf{static setup}, 
where training and test sets are randomly split and
tend to exhibit similar data distributions~\cite{glandt-etal-2021-stance, hansen-etal-2021-automatic, DBLP:conf/aaai/MathewSYBG021}.
It is thus incapable of reflecting the realistic  application scenarios --- a model should usually learn to tackle the data created after it is trained while the evolving features would continuously shift the data distributions.

Language learning with distribution shift 
(a.k.a., OOD, short for out-of-distribution) has drawn a growing attention in the NLP community~\cite{DBLP:journals/corr/abs-2108-13624, arora-etal-2021-types}.
Most previous work focuses on OOD in different domains~\cite{DBLP:conf/icml/MuandetBS13, DBLP:journals/corr/GaninUAGLLML15} and studies how to learn generalizable cross-domain features. 
Here we experiment OOD in the dynamic environment --- whose time-sensitive nature renders the data evolution to occur progressively and continuously;
whereas most prior empirical studies discuss OOD across domains
and hence focus on the relatively discrete shifts from the source to target domains~\cite{DBLP:conf/nips/VolpiNSDMS18, DBLP:conf/icml/KruegerCJ0BZPC21}.

To further examine NLU adaption to time evolution
(henceforth \textbf{time-adaptive learning}), we exploit a small set of unlabeled data posted after a model is trained (henceforth \textbf{trans-data}) and investigate its potential in mitigating the time-shaped feature gap.
For methodology, we start with the existing solutions in unsupervised domain adaption (UDA)~\cite{ramponi-plank-2020-neural} and employ two popular baselines in this line, one is feature-centric based on auto-encoding (specifically VAE) and the other data-centric pseudo-labeling (PL).
Furthermore, a joint-training framework is  explored to study their coupled effects in fighting against the possible performance deterioration over time.

The experiments are based on three 
trendy social media tasks about the detection of COVID-19 stance ~\cite{glandt-etal-2021-stance}, fake news ~\cite{hansen-etal-2021-automatic}, and hate speech ~\cite{DBLP:conf/aaai/MathewSYBG021} with the benchmark data from Twitter.
We also gather a new corpus for hashtag prediction to broaden our scope to noisy user-generated labels tremendous on social media.\footnote{Hashtags are tagged by the author of a post to indicate its topic label and start with an hash ``\#'', e.g., ``\#COVID19''. } 
Dynamic setup is adopted and models are tested on multiple datasets varying in the time gap to the training data to quantify the model sensitivity to the time evolution.

In the main results, the performance of all models are gradually worse in general over time.
It implies dynamic social media environment may universally and negatively affect the NLU effectiveness.
With some trans-data, both VAE and PL can helpfully tackle dynamicity and the their joint framework achieves the best results consistently over time.
We then analyze the effects of trans-data scale and create time and find both PL and VAE might benefit from trans-data with larger scales and smaller time gap to the training data. 
At last, 
case studies
interpret how VAE and PL collaboratively handle the dynamic environments. 

To conclude, \emph{we present the first empirical study, to the best of our knowledge, on the universal effects of dynamic social media environment on NLU, and provide insights to when and how UDA methods help advance  model robustness over time. } 

%% file: figures/intro-case.tex
\begin{figure*}[t]\small
\centering
\begin{tabular}{p{0.5cm}p{9cm}c}
\cline{1-2}
\textbf{Time}&\textbf{Topics 
}&\multirow{6}{*}{\includegraphics[width=.33\textwidth]{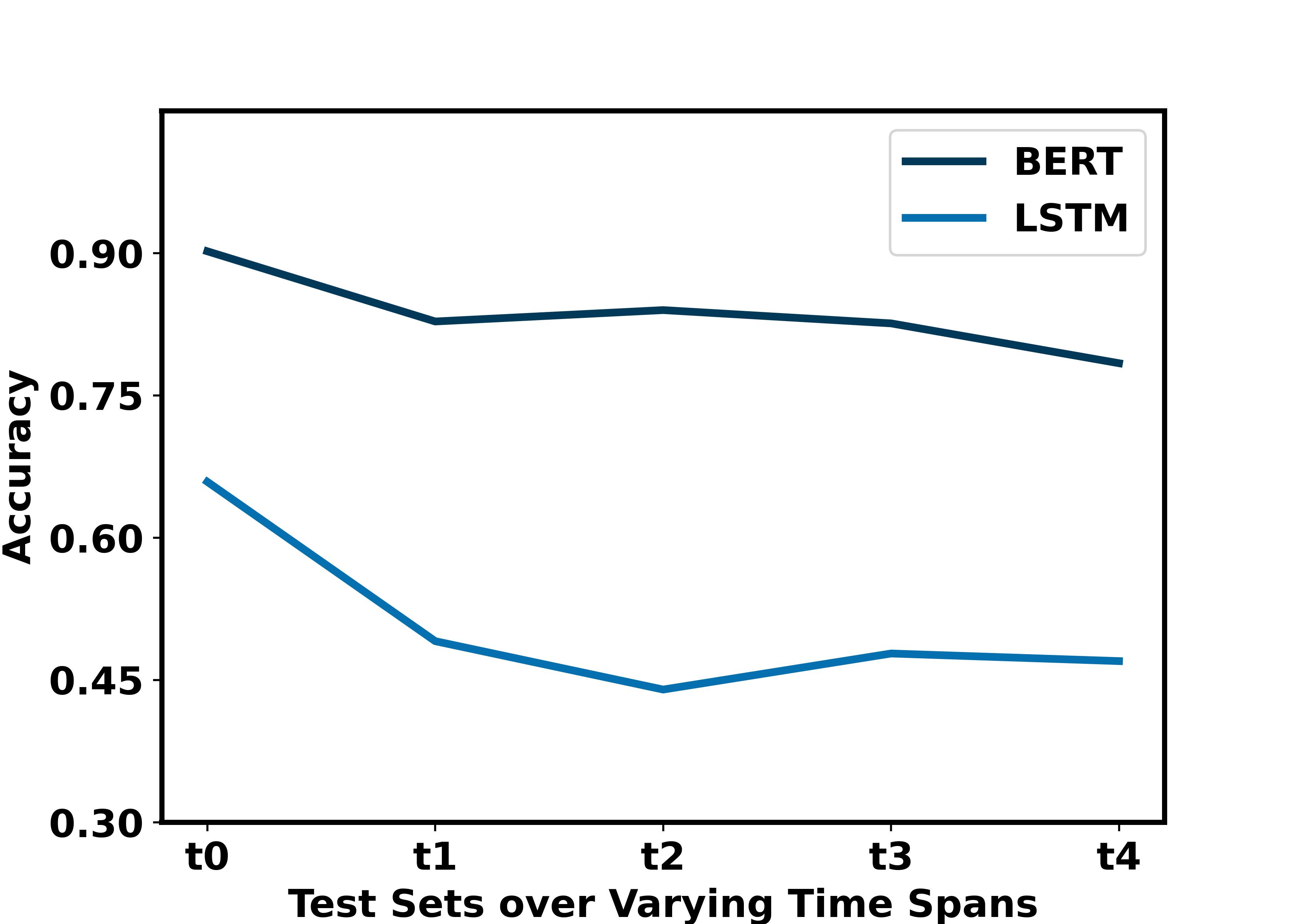}}\\

\cline{1-2}

$t_{0}$&Scientist Doctor, Covid, Corona Virus, No Mask, Return To Work, Govern, Nose, Mouth, Dread& \\

$t_{1}$&Real Patriots Wear Mask, Failed Lock Down, 2020 US Election, Save, Immune Compromise, Covid19 Outbreak, Schools Wear a Mask,  Corona Virus Pakistan, Corona Virus Canada& \\

$t_{2}$&Symptom, Temperature, Lock Down, Panic Buy, Cough, I m With Fauci, Trump Kills Us, Mask Up& \\

$t_{3}$&Therapy, Inject, Wear Your Mask, Trump Is A National Disgrace, Trump Land Slid, Surgeon, Covid 2019 India, Wear You Masks Dont Work& \\

$t_{4}$&Red State, Blue State, Trump Lies Americans, Lying Trump, Trump Melt Down, End Lock Down& \\

\cline{1-2}

\end{tabular}
\vspace{-0.5em}
\caption{
Results from the Twitter stance detection dataset for COVID-19 topics~\cite{glandt-etal-2021-stance}.
$t_0$ refers to the time span of the earliest 40\% tweets and the rest are equally split into 4 segments in the chronological order corresponding to $t_1$, $t_2$, $t_3$, and $t_4$, respectively.
Latent topics from $t_0$ to $t_4$ are shown on the left and topic words are learned by VAE. 
Stance detection results over time are shown on the right, where x-axis indicates test sets from $t_0$ to $t_4$ and y-axis the prediction accuracy.
LSTM results are displayed in the light blue line and BERT dark blue.
}
\vspace{-0.5em}
\label{fig:intro-case}
\vspace{-0.5em}
\end{figure*}

%% file: sections/related-work.tex
\section{Related Work}
This paper is in line with previous work for the out-of-distribution (OOD) issue, aiming to mitigate the distribution gap between training and test data
\cite{DBLP:conf/iclr/XieKJKML21, DBLP:journals/corr/abs-2108-13624, DBLP:journals/corr/abs-2101-05303}.
Most prior OOD studies experiment on domain gaps which tend to exhibit intermittent change, while that shaped by time usually happen step by step and hence forms a successive process.
Limited attention has been paid to examine NLU models' practical and general performance in handling evolving social media environment, while our empirical study is an initiate to fill in the gap.

In previous OOD work, various domain adaptation methods are explored~\cite{chu-wang-2018-survey, ramesh-kashyap-etal-2021-domain}, e.g., adversarial learning~\cite{DBLP:journals/corr/abs-2004-08994},  pre-training~\cite{hendrycks-etal-2020-pretrained, goyal-durrett-2020-evaluating, kong-etal-2020-calibrated}, and data augmentation~\cite{DBLP:conf/pkdd/ChenLWLJ21}. 
Some of them require labeled data from both source and target domains 
~\cite{arora-etal-2021-types} to learn cross-domain features. 
It is however infeasible in our time-shaped OOD scenarios
because of the difficulties to continuously label data.  


Our baseline solutions are inspired by existing methods in unsupervised domain adaption (UDA), employing labeled source data and unlabeled target data for model training.
Popular UDA baselines mostly fall into feature-centric and data-centric categories \cite{ramponi-plank-2020-neural}.
The former explores implicit clusters to bridge semantic features across domains~\cite{gururangan-etal-2019-variational}, while the latter transfers knowledge gained from the source to target via self-training~\cite{axelrod-etal-2011-domain}.
In our experiments, VAE and pseudo-labeling (PL) are popular baselines respectively selected to represent feature- and data-centric UDA, whereas their individual and collaborative performance in advancing NLU robustness in dynamicity have never been studied before and will be explored here.

This work is also inspired by the previous studies applying
a dynamic setup on certain social media tasks, 
such as content recommendation \cite{zeng-etal-2020-dynamic,zhang-etal-2021-howyoutagtweets}.
Based on their efforts, 
we take a step further to broadly examine various classification tasks in order to draw a more general conclusion on how dynamicity affects NLU.

%% file: sections/framework.tex
\section{Time-Adaptive Learning Baselines 
}
\input{figures/model}

Here we discuss  time-adaptive learning baselines and how we leverage them with social media classification  in dynamicity.
We will start with the classification overflow, followed by the introduction to VAE- and PL-based baselines ($\S$\ref{ssec:model:time}) and how they can collaboratively work via joint training ($\S$\ref{ssec:model:losses}).


\paragraph{Classification Overflow.}
The NLU in a dynamic setup will be examined on multiple tasks, all formulated as post-level single-label classification. The input is a social media post $s$ and output a label $l$ specified by a task. 
Here we assume the availability of two data types: 
(1) posts with gold-standard labels created in the past  (henceforth \textbf{history labeled data}), which can be employed to train a supervised classifier;
(2) posts created after the classifier is trained and without labels (i.e., trans-data).

For classification, following the advanced practice \cite{devlin-etal-2019-bert}, the  representation for language  understanding will be built in a pre-trained BERT encoder, where we feed in the input $s$ and obtain a latent vector $\mathbf{b}_s$ as the post embedding.
%
%


At the output, the learned classification features are mapped to a specific label $\hat{y}_{s}$ with the formula:


\begin{equation}\small
    \hat{y}_{s}=\textit{f}_{out}(\mathbf{W}_{out}\cdot  \mathbf{r}_{s}+\mathbf{b}_{out})
\end{equation}

\noindent $\textit{f}_{out}(\cdot)$ is the activation function for classification output (e.g., softmax). $\mathbf{W}_{out}$ and $\mathbf{b}_{out}$ are learnable parameters for training.
$\mathbf{r}_s$  couples the BERT-encoded latent semantics ($\mathbf{b}_s$) and the implicit cross-time features gained by VAE (as a feature-centric UDA) via a multi-layer perceptron (MLP):

\begin{equation}\small\label{eq:prediction}
    \mathbf{r}_{s}=
    f_{MLP}(\mathbf{W}_{MLP}[\mathbf{b}_s;\mathbf{z}_s]+\mathbf{b}_{MLP})
\end{equation}

\noindent $\mathbf{z}_s$ is yielded by VAE through the  process to be later discussed in $\S\ref{ssec:model:time}$. 
$f_{MLP}(\cdot)$ is the ReLU activation function. $\mathbf{W}_{MLP}$ and $\mathbf{b}_{MLP}$ are both learnable.






\subsection{Time-Adaptive Learning Methods}\label{ssec:model:time}

Here we first describe two UDA baselines to be tailor-made to our setup, feature-centric VAE and data-centric PL.
Then we discuss how we integrate them to explore their collaborative effects.

\paragraph{VAE.} 
The potential of VAE to deal with dynamicity comes from its capability of clustering posts exhibit similar word statistics and forming latent topics to reflect their shared discussion point.
Therefore, the intra-cluster content, though varying in the generation time,  reflects the implicit semantic consistency throughout the time and enables the learning of underlying past-to-future connection.

In the following, we detail how VAE is applied to tackle dynamicity.
First, both the history data (has labels and used to train the classifier) and unlabeled trans-data (created after the classifier is trained) are gathered together to form a corpus. 
Then, VAE is employed to explore the global word statistics throughout the entire the corpus via clustering.  
Here we implement VAE following the widely-used design from
\citet{DBLP:conf/icml/MiaoGB17}, where a post $s$ is fed into the auto-encoder in the bag-of-word (BoW) vector form, denoted as $\mathbf{v}_{s}$, for easier statistical measure. 

Given the BoW input $\mathbf{v}_{s}$, the clustering is conducted through auto-encoding, which contains an encoding process to map $\mathbf{v}_{s}$ into a latent topic indicator $\mathbf{z}_s$, followed by a decoding to rebuild $\mathbf{v}_{s}$ conditioned on the topic ($\mathbf{z}_s$).
$\mathbf{z}_s$ is a $K$ dimensional vector; each entry reflects the chance $s$ should be clustered into a certain topic and $K$ is a hyper-parameter indicating the total topic number in the corpus. Below presents the concrete steps.

For encoding, $v_{s}$ is embedded into the latent topic space to generate $\mathbf{z}_s$ via Gaussian sampling, where the mean $\mu$ and standard deviation $\sigma$ are learned with the following formula:

\begin{equation}\small
    \mu=f_\mu(f_e(\mathbf{v}_s)), log\,\sigma=f_\sigma(f_e(\mathbf{v}_s))
\end{equation}

\noindent $f_*(\cdot)$ is a ReLU-activated neural perceptron. 
Then $\mathbf{z}_s$ is drawn from the normal distribution below:

\begin{equation}\small
    \mathbf{z}_s=\mathcal{N}(\mu, \sigma)
\end{equation}

It is later transformed to a distributional vector via softmax to yield $\theta_s$, representing the topic mixture of $s$. It initiated decoding step to re-construct $\mathbf{v}_s$ 
by predicting $\hat{\mathbf{v}}_s$ below:


\begin{equation}\small
    \hat{\mathbf{v}}_s=\textit{softmax}(f_\phi(\theta_s))
\end{equation}

\noindent $f_\phi(\cdot)$ is another ReLU-activated perceptron mapping information in topic space back to the BoW. 
The weights of $f_\phi(\cdot)$ (after softmax normalization) are employed to represent the topic-word distributions and the latent topic vector $\mathbf{z}_s$ (with cross-time views gained in clustering) can be engaged in classification (Eq.  \ref{eq:prediction}) to advance robustness over time.

\paragraph{PL.} As discussed above, VAE, as a feature-centric baseline, tends to explore shared features across time to mitigate the OOD.
However, supervision from history data labels, is not explicitly leveraged in the modeling of cross-time features.
The potential to further engage data labels can be explored in the data-centric baselines, among which pseudo-labeling (PL) demonstrates simple-yet-effective in previous OOD   experiments in domain adaption~\cite{axelrod-etal-2011-domain}.

Here, the priorly trained classifier (with the labeled data from the past) is first pseudo-label the trans-data without labels.
Then, pseudo-labels later engage in the task supervision and work together with the labeled history data to continuously train the classifier.
In this way, the models may self-learn how to adapt the knowledge gained from the past supervision to the future scenarios through the pseudo-labeled trans-data, and therefore enable a ``dynamic supervision'' in the time evolution.


\paragraph{Integrating VAE and PL.}
We have shown the potential of VAE and PL baselines in tackling dynamicity, where the former gains a time-adaptive view through features and the latter labels. 
It is thus interesting to explore how they can collaborate with each other to enable a better robustness. 

We therefore design a integrated framework to couple the effects of both  VAE-learned features and PL-predicted trans-data.
The framework architecture is shown in Figure \ref{fig:model}.
As can be seen, ``time-adaptive features'' are formed via concatenating the BERT-encoded post embedding ($\mathbf{b}_s$) and VAE-encoded cross-time features ($\mathbf{z}_s$), which are fed into MLP for classification (Eq. \ref{eq:prediction}).
Meanwhile, the priorly-trained classifier is employed to predict pseudo labels on trans-data, which joint hands with the labeled history data to form the ``time-adaptive data''.
The classifier with ``time-adaptive features'' is then continuously trained on the ``time-adaptive data'' to gain the joint advances of VAE and PL.


\subsection{Joint Training}
\label{ssec:model:losses}
We have discussed how to leverage the pre-trained VAE features in an integrated PL+VAE (PV) framework (in Figure \ref{fig:model}).
We are further interested in whether we can jointly explore the classification training of PV and the unsupervised learning of VAE, which may enable VAE to learn cross-time features in aware of the PL-enhanced view from PV. The joint training is conducted via optimizing the training losses of PV classification and VAE. 




For PV, the loss is based on the cross-entropy.
The training set $\tau$ consists of post-label pairs $(s,l)$ from both history data with gold-standard labels and trans-data with pseudo-labels. 
The PV training loss $\mathcal{L}_{p}$ is defined as:


\begin{equation}\small\label{eq:engage-loss}
    \mathcal{L}_{p}=-\sum_{(s,l)\in \tau}(y_{s,l} \textit{log} (\hat{y}_{s,l})+(1-y_{s,l})\textit{log}(1-\hat{y}_{s,l}))
\end{equation}

For the VAE loss,  
we use variational inference to approximate a posterior distribution over a post's 
topic $\mathbf{z}_s$ given word statistics observed in its context.
Here we formulate the VAE loss $\mathcal{L}_{vae}$  as:

\begin{equation}\small\label{eq:vae-loss}
    \mathcal{L}_{vae}=D_{KL}(p(\mathbf{z}_s)\,||\,q(\mathbf{z}_s|s))-\mathbf{E}_{p(\mathbf{z}_s)}[p(h|\mathbf{z}_s)]
\end{equation}

\noindent Here $D_{KL}(\cdot)$ is the Kullback-Leibler divergence
loss and $\mathbf{E}_{*}[\cdot]$ measures the VAE reconstruction. \\


PV and VAE losses are then added with weights to produce the joint-training loss:


\begin{equation}\small\label{eq:overall-loss}
    \mathcal{L}=\mathcal{L}_{p}+\mu \cdot \mathcal{L}_{vae}
\end{equation}

\noindent where $\mu$  trades-off the PV and VAE effects. In this way, parameters of VAE and PV are updated together to jointly tackle  time-adaptive learning.


%% file: figures/model.tex
\begin{figure}[tb]
    \centering \includegraphics[width=.4\textwidth]{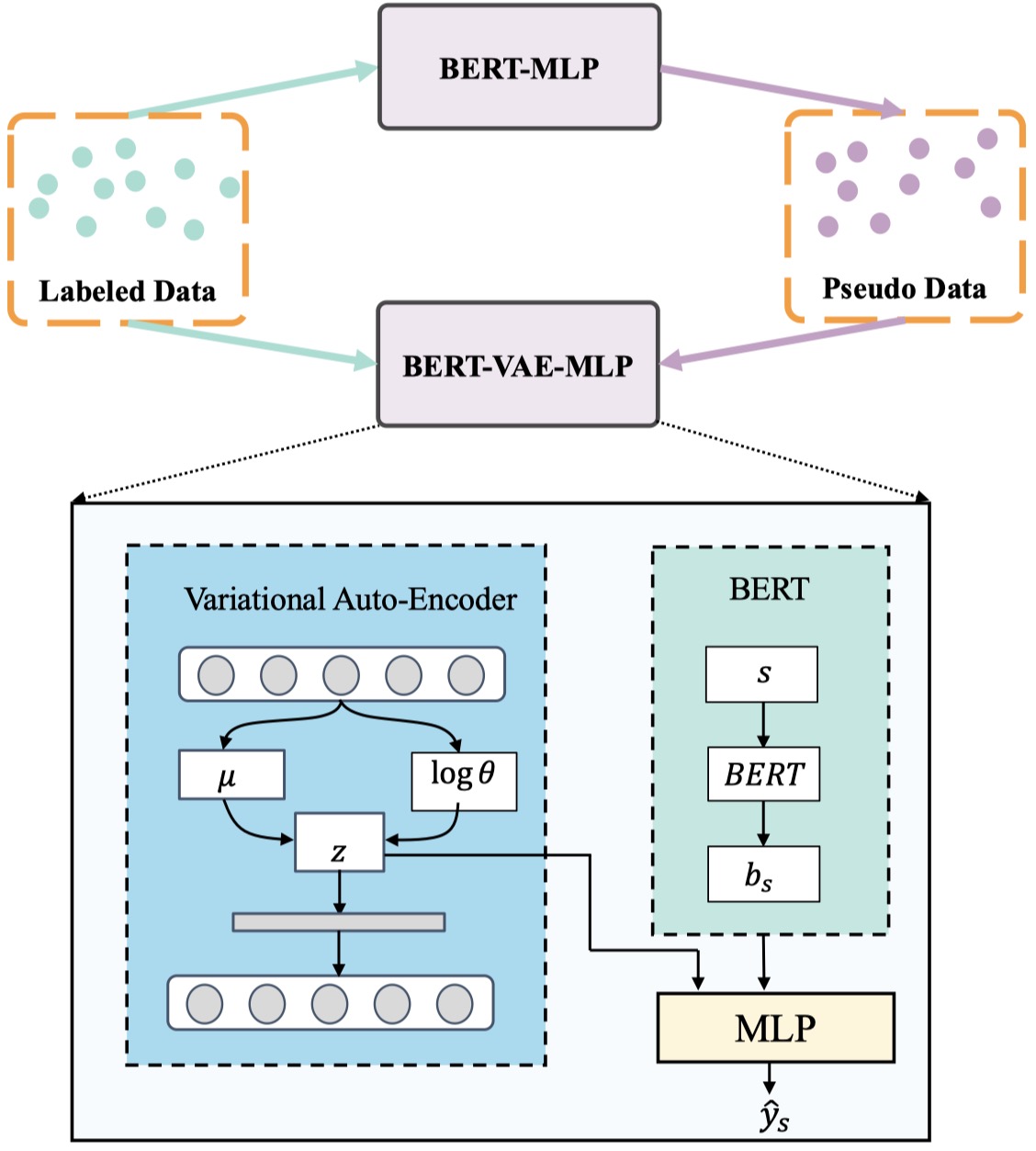}
    \caption{  \label{fig:model}
    Our integrated framework of VAE and PL. VAE-learned features ($\mathbf{z}_s$) are injected into MLP with BERT-encoded $\mathbf{b}_s$ (indicated as BERT-VAE-MLP). 
    PL-predicted trans-data (pseudo data) and the labeled data from the past are both used to train  BERT-VAE-MLP.
    }
    \vspace{-1em}
\end{figure}

%% file: sections/datasets-setup.tex
\section{Datasets and Experimental Setup}



\input{tables/dataset}

\subsection{Experimental Datasets}\label{ssec:exp:dataset}

We experiment models' performance in dynamic on four popular tasks for the detection of stance, fake news, hate speech, and hashtags.
Their corresponding datasets are all from Twitter and each sample is a tweet with a classification label.
For the tweets with missing information, e.g., the time stamp, we adopt the Twitter API for a recover.\footnote{\url{https://developer.twitter.com/en/docs/twitter-api}}

The \textsc{Stance} detection dataset contains tweets with annotated stance about various COVID-19 topics \cite{glandt-etal-2021-stance}.
For a straightforward comparison to other tasks, we employ models to only predict the stance labels of ``favor'', ``against'', and ``none'', regardless of the topics. 



For the \textsc{Fake-News}, we employ the Twitter dataset from
\citet{hansen-etal-2021-automatic}.
Due to the imbalanced labels in small-scaled data, we only consider the coarse-grained classification of whether the a tweet stated ``true'' or ``false'' information. 


The \textsc{Hate-Speech} data is released by~\citet{DBLP:conf/aaai/MathewSYBG021} with binary labels indicating whether or not the hate speech exists in a tweet.


The above three datasets are from publicly available benchmarks with relatively clean annotations. 
However, many social media applications are built upon noisy user-generated labels \cite{wang-etal-2019-topic-aware,zhang-etal-2021-howyoutagtweets}.
Here we take hashtag prediction as an example and newly construct a dataset, namely \textsc{Hashtag}, where the input is a tweet and output a hashtag tagged by its author.

For the data collection, we first follow \citet{nguyen-etal-2020-bertweet} to gather large-scale tweets and those posted during September - December, 2011 were selected, to roughly allow a balanced year coverage across different datasets.
Following previous practice \cite{zeng-etal-2018-topic}, the top 50 hashtags with the highest frequency are selected to be the labels. 

Table \ref{tab:accents} shows an overview of the dataset, all reflecting varying statistics and characteristics. 
It allows us to experiment with various scenarios.



\subsection{Dynamic Setup and Data Analysis}\label{ssec:exp:setup}
\input{figures/relevance}

As discussed above, instead of the random dataset split, we adopt a more realistic dynamic setup.
A specific cut is applied over the chronologically ordered tweets, where the earliest samples (their posting time span in $t_0$ period) are for training, and the rest for test.
For datasets with imbalanced distribution over time, the cut is customized to avoid too small test sets.
Specifically, the cut over \textsc{Stance} results in a 4:6 split ($t_0$ corresponds to the earliest 40\% tweets) while that over \textsc{Fake-News} and \textsc{Hate-Speech} is 6:4.
For \textsc{Hashtag} with much more balanced time distribution, we employ an absolute cut and take the September data for $t_0$. 

To further test models in varying degrees of data freshness, the tweets posted after $t_0$ are split into four slices (with equal size and ordered by time), corresponding to $t_1$, $t_2$, $t_3$, and $t_4$. 
Tweets posted in each period form a test set thereby exhibits a gradually larger time gap to the training data.

Finally, for a comparison with a common-used static step, the $t_0$ data is randomly split into 80\% training, 10\% validation, and 10\% test, where $t_0$ test set may distribute similarly to the training. 

We preliminarily analyze test data in \textsc{Hashtag}, the largest dataset, and examine the vocabulary overlap 
between pair-wise periods following~\citet{gururangan-etal-2020-dont}. 
Results are shown in Figure \ref{fig:relevance} and longer time gap in general exhibits a smaller overlap.
It challenges a priorly-trained model with more unseen features, while  trans-data, statistically closer to the test, helpfully mitigate the gap.

\subsection{Comparisons and Model Settings}
\label{ssec:exp:baseline}

In baseline setups, we first considered supervised classifiers LSTM (initialized with GloVe embeddings~\cite{pennington-etal-2014-glove}) and BERT (pre-trained BERTweet \cite{nguyen-etal-2020-bertweet}). 
The hidden layer size of MLP was set to 2048. 
For time-adaptive baselines, VAE was implemented based on
Vampire~
\cite{gururangan-etal-2019-variational} with topic number $K$ set to 50.
The hyperparameters were set via grid search on validation with 40 epochs.

For PV, the integrated PL+VAE time-adaptive learning framework, we compare two training strategies: S-PV (PV with priorly-trained VAE) and J-PV (PV and VAE are jointly trained following description in $\S$\ref{ssec:model:losses})
In J-PV, PV and VAE loss are added with the weight $\mu$ set to 1e-2 (Eq. \ref{eq:overall-loss}).

For time-adaptive learning evaluation, inductive learning is set up for $t_0$ - $t_3$, assuming the availability of unlabeled test data as the trans-data; while for $t_4$, we follow transductive learning setup to employ $t_1$ - $t_3$ data as the trans-data and test on $t_4$, which reflects the realistic scenarios where trans-data should be gathered before the application.



%% file: tables/dataset.tex
\begin{table}[t]\small
\centering
\begin{tabular}{lrr}
\hline
\textbf{Dataset}&\textbf{Scale} & \textbf{Time Span}\\
\hline
\textsc{Stance}&7,122&2020/Feb-2020/Aug\\
\textsc{Fake-News}&8,847&2006/May-2018/Nov\\
\textsc{Hate-Speech}&8,773&2018/Oct-2019/Oct\\
\textsc{Hashtag}&30,018&2011/Sep-2011/Dec\\
\hline
\end{tabular}
\vspace{-0.5em}
\caption{
Statistics of the four experimental datasets.
}
\vspace{-1em}
\label{tab:accents}
\end{table}

%% file: figures/relevance.tex
\begin{figure}[t]
    \centering
    \includegraphics[width=.3\textwidth]{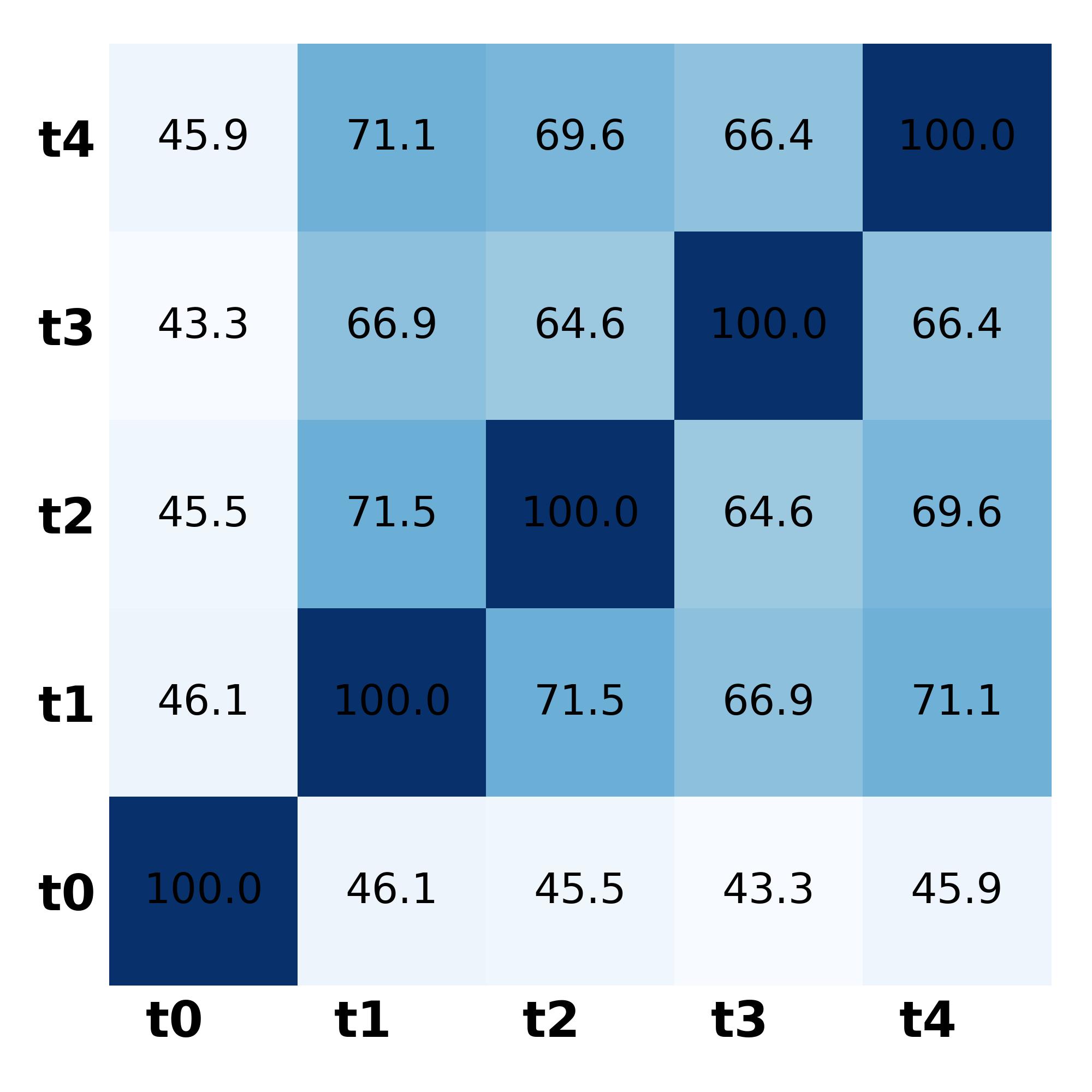}
        \vspace{-0.5em}
    \caption{
    Pairwise vocabulary overlap (\%) within   $t_{0}$ - $t_{4}$ test sets.
    The vocabulary gathers the
    top 1K most frequent words (excluding stop words)
    from each set.
    }
    \vspace{-0.5em}
    \label{fig:relevance}
    \vspace{-1em}
\end{figure}

%% file: sections/results-analysis.tex
\section{Results and Analysis}

In this section, we first present the comparison results test over $t_0$ to $t_4$
($\S$\ref{ssec:main-results}). 
Then we discuss more about trans-data via
quantifying the 
data scale 
and freshness 
($\S$\ref{ssec:analysis}). 
At last, $\S$\ref{ssec:case-study} shows a case study to interpret how VAE and PL work together to enable model adaption to the feature change.

\subsection{Main Results in Dynamic Setup}
\label{ssec:main-results}

Table~\ref{tab:main results} shows the comparison results over time and the following observations can be drawn.

\input{tables/main-results}

First, all models perform the best on $t_0$ and after that roughly exhibit a gradually worse accuracy over time.
It demonstrates the evolving environment would indeed negatively affect the social media classification, which may universally happen to various tasks;
furthermore, such negative effects may be enlarged as time goes by.
The possible reason lies in the ever-changing features, continuously providing something a model, trained on the history data, has never learned (as we induced from Figure \ref{fig:relevance}).
Between BERT and LSTM, the former seems to exhibit a better over-time robustness, probably because the generic NLU capability gained through large-scale pre-training might offer some help.
And BERT, unsurprisingly, handles $t_0$ test (static setup) excellently; its results over $t_1$ - $t_4$ (dynamic setup) are still compromised, calling for our community's collaborative efforts on a time-adaptive NLU.



Second, both VAE and PL, in general, help mitigate the performance drop over time, whereas they are superior in different scenarios.
VAE seems to better perform in inductive learning ($t_0$-$t_3$) while PL works better in transductive learning ($t_4$). 
It might be because VAE explores feature-level connection across time, while the inductive data, though without labels, may allow models to better fit their feature space to the test.
On the contrary, PL gains time-adaptive knowledge through the pseudo labels and is therefore less sensitive to whether the trans-data is inductive or transductive.



Third, our proposed J-PV, outperforms its variant S-PV and champion three tasks except \textsc{Hashtag} on transductive evaluation over $t_4$.
For \textsc{Hashtag}, it also obtains the second best accuracy though outperformed by S-PV.
The possible reason is that \textsc{Hashtag} requires the learning of noisy and diversely distributed user-generated labels ($\S$\ref{ssec:exp:dataset}). 
It may somehow challenges the joint training with unsupervised VAE;
instead, the pre-trained VAE in S-PV enables models to priorly gain a global overview of the dataset exhibits various topics, and hence result in the best test accuracy on $t_1$ - $t_4$ (all dynamic setups).
However, for the other three datasets, S-PV fails to well coordinate PL and VAE (sometimes even worse than either of them).

From the last two points above, we further learn that PL and VAE, skilled differently, may potentially enable positive collaborative effects, whereas a well-designed method is also needed to better couple their advantages.
A simple-yet-effective joint training has shown some initially promising results, which can be carried on in future work.



\subsection{
Further Discussions on Trans-Data}
\label{ssec:analysis}

We have shown, in $\S$\ref{ssec:main-results}, the promising results of time-adaptive learning with some trans-data.
As a pilot experiment here, it is assumed that trans-data is available (set up with tweets from the same dataset and created after $t_0$).
However, in the real world, how to automatically gather and select trans-data might become another research question.

To provide some insight to trans-data study, we further discuss how the trans-data scales and freshness (time gap to the training) affect transductive test results on $t_4$.
In the following, \textsc{Hate-Speech} is taken as an example to discuss here while similar observations are drawn from other datasets. 




\input{figures/ana-scale}
\vspace{-0.5em}

\paragraph{Effects of Data Scale.}
Here we randomly shuffle the trans-data and feed varying it in varying scales to VAE and PL baselines.
The results are shown in Figure \ref{fig:ana-scale} and we observe both PL and VAE, in general, benefit from relatively more trans-data. 

Compared to PL, VAE seems to flatten its trends after 1/3 data is given.
Also, 
interesting, VAE helps BERT perform better even without trans-data, probably because its latent clusters enable better semantic learning from noisy tweets with sparse context.

For PL, it offers no benefit with no trans-data whereas it peaks its accuracy at 2/3 data.
It is possibly because, 
PL leverages labeled data from the past and unlabeled trans-data, to capture time-adaptive skills; additional data may allow the model to see more data while also higher the risk of being affected by pseudo-labeling errors.




To further probe how sensitive PL is to the pseudo-labeling accuracy, we quantify PL's test results, in Figure \ref{fig:ana-label}, through the learning with varying accuracy groups of pseudo-labeled samples.
We observe the wrongly-labeled negative samples result in worse accuracy compared to PL learning without trans-data.
It shows the errors in pseudo-labeling, unavoidable in self-training, would indeed risk the PL effectiveness.
However, when working together with the  positive samples, PL achieves promising results, only slightly worse than the upper-bound results (with correctly-labeled data only).

\input{figures/ana-label}


\paragraph{Effects of Data Freshness.} 
We then analyze the effects of trans-data freshness (time gap to the training data), where $t_1$, $t_2$, and $t_3$ data take turns to be used to train PL and VAE.
As can be seen, for both PL and VAE, trans-data, regardless of its freshness, enables obvious performance gain over the BERT ablation trained without  trans-data.

It is also observed that PL seems to benefit more from trans-data with relatively smaller time gap to the training. The reason is such trans-data enables easier pseudo-labeling, resulting in more positive samples and better overall performance (as we discussed in Figure \ref{fig:ana-label}).
For VAE, the training with $t_1$ or $t_3$ data is more effective than $t_2$.
It is possibly because the trans-data, temporally closer to either training or test data, would allow a model to better connect its embedded semantics with  what is seen in training or test, either would help signal the past-to-future change to shape cross-time features.




\input{figures/ana-time}

\subsection{Case Study}
\label{ssec:case-study}

We have shown in Table \ref{tab:main results} a jointly-trained VAE and PL (J-PV) can effectively deal with changing features.
Here we take cases from the \textsc{Stance} to interpret model output and discuss how it works.
Recall that the dataset contains tweets reflect user stances on COVID-19 topics and we have observed a change of users' focuses from $t_0$ to $t_4$ through the VAE clustered topics (Figure \ref{fig:intro-case}).

Here we specifically examine the cases related to Dr. \textit{Anthony Fauci}, to whom we notice an obvious increasing supporting rate, with 56\%, 54\%, 89\%, and 74\% tweets containing ``fauci'' and in favor of him through $t_0$ to $t_3$. 
Our J-PV model, via collaborating PL and VAE over $t_1$-to-$t_3$ trans-data, is able to better predict the ``favor'' stance to Fauci-tweets compared to the BERT ablation trained on $t_0$ only. 

We qualitative analyze the reasons and draw three samples $S_1$, $S_2$, and $S_3$, respectively from $t_1$, $t_2$, and $t_3$.
Their self-attention weights are visualized in Figure \ref{fig:heatmap}, where the model obviously highlights ``CoVirus'' and  ``Trump''-words together with ``Fauci''-words.
It is possibly because, when J-PV self-trains on trans-data, it tries to align the stance learned from the dominant ``CoVirus''-tweets in $t_0$ (Figure \ref{fig:intro-case}) via pseudo-labeling.
Meanwhile, from $t_1$ to $t_3$, VAE detects trendy topics in ``CoVirus'', ``Trump'', and ``Fauci'', as their their related words frequently co-occur.  
Further, the ``usually-against'' stance gained for ``Trump'' might in return  indicates the ``favor'' in ``Fauci'', when they two are both discussed.
Interestingly, we do observe a higher ``for-Fauci'' stance rate in tweets mentioning both Trump and Fauci, from 35\% in $t_0$ to $68\%$ in $t_2$.

\input{figures/heatmap}

%% file: tables/main-results.tex
\begin{table}[htb]\small
\centering
\begin{tabular}{p{0.4cm}p{0.7cm}p{0.7cm}p{0.7cm}p{0.7cm}p{0.75cm}p{0.7cm}}
\hline
\textbf{Time} &\textsc{LSTM}& \textsc{BERT} & \textsc{VAE} & \textsc{PL} & \textsc{S-PV}& \textsc{J-PV}\\

\hline
\multicolumn{7}{c}{\textsc{Stance}}\\
\hline
$t_{0}$&0.659&0.902&0.884&0.899&0.906&0.891\\
$t_{1}$&0.491&0.828&0.840&0.831&0.830&0.839\\
$t_{2}$&0.440&0.840&0.839&0.840&0.817&0.850\\
$t_{3}$&0.478&0.826&0.830&0.826&0.821&0.840\\
$t_{4}$&0.470&0.784&0.794&0.799&0.785&\textbf{0.804}\\

\hline
\multicolumn{7}{c}{\textsc{Fake-News}}\\
\hline
$t_{0}$&0.576&0.629&0.633&0.617&0.600&0.596\\
$t_{1}$&0.557&0.615&0.634&0.607&0.607&0.605\\
$t_{2}$&0.562&0.656&0.654&0.653&0.648&0.652\\
$t_{3}$&0.550&0.622&0.620&0.630&0.631&0.631\\
$t_{4}$&0.514&0.636&0.659&0.669&0.676&\textbf{0.685}\\

\hline
\multicolumn{7}{c}{\textsc{Hate-Speech}}\\
\hline

$t_{0}$&0.683&0.679&0.698&0.676&0.687&0.677\\
$t_{1}$&0.772&0.783&0.771&0.779&0.762&0.745\\
$t_{2}$&0.755&0.768&0.779&0.769&0.772&0.771\\
$t_{3}$&0.621&0.648&0.640&0.649&0.647&0.661\\
$t_{4}$&0.539&0.612&0.616&0.622&0.613&\textbf{0.632}\\
\hline
\multicolumn{7}{c}{\textsc{Hashtag}}\\
\hline

$t_{0}$&0.716&0.771&0.781&0.646&0.679&0.664\\
$t_{1}$&0.563&0.602&0.619&0.622&0.642&0.635\\
$t_{2}$&0.540&0.612&0.612&0.654&0.671&0.664\\
$t_{3}$&0.520&0.648&0.653&0.708&0.722&0.717\\
$t_{4}$&0.531&0.649&0.654&0.684&\textbf{0.704}&0.689\\

\hline

\end{tabular}
\vspace{-0.5em}
\caption{\label{tab:main results}
The classification accuracy over time (through $t_0$-$t_4$ test), where higher scores indicate better results.
J-PV champions transductive test on $t_4$ over three datasets, though slightly outperformed by S-PV on \textsc{Hashtag}.
}
\vspace{-1em}

\end{table}

%% file: figures/ana-scale.tex
\begin{figure}[t]
    \centering
    \includegraphics[width=.35\textwidth]{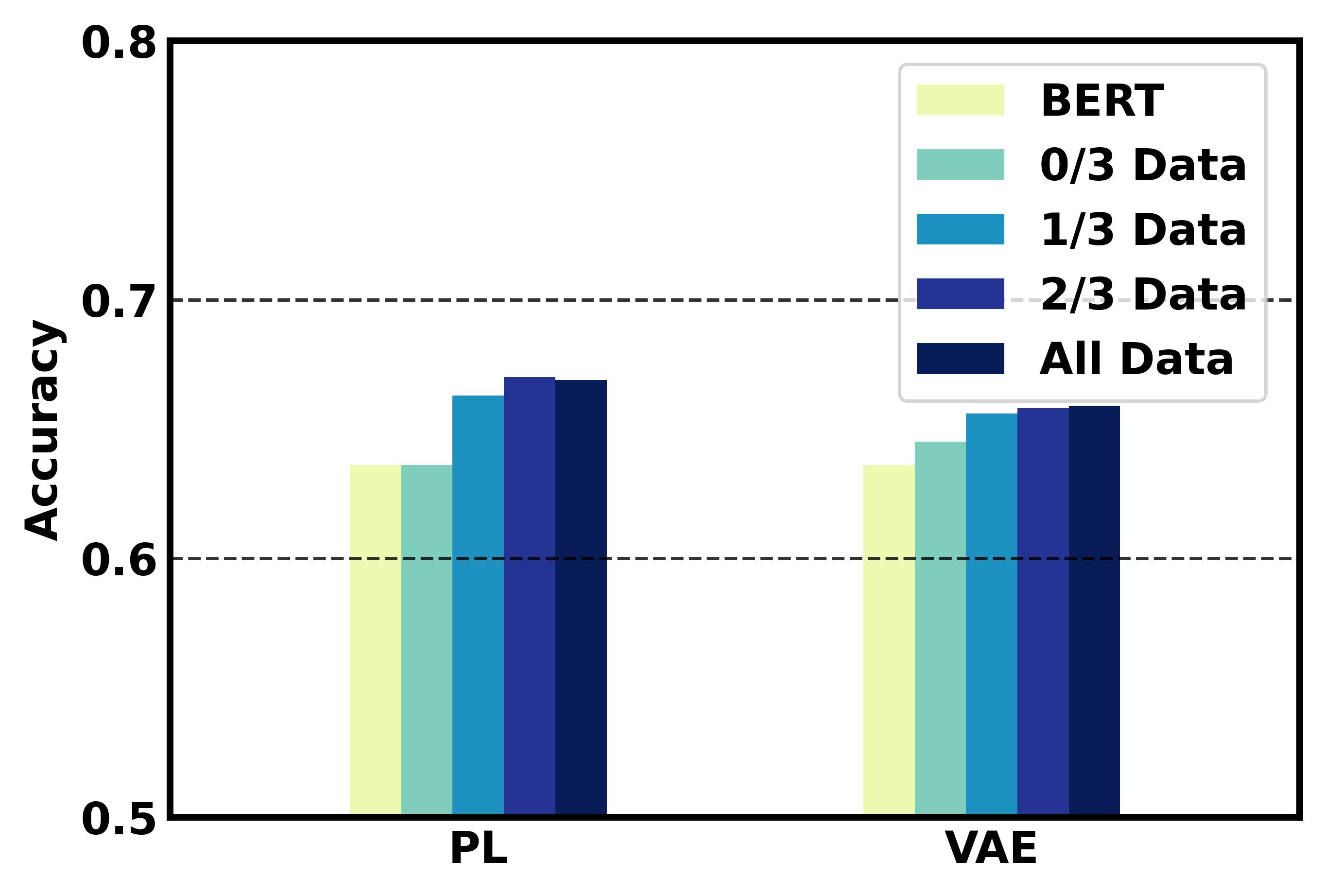}
    \vspace{-0.5em}
    \caption{Test accuracy on $t_{4}$ (y-axis).
    The left bar group refer to PL results and right VAE.
    In each group, from left to right shows BERT (leftmost) and that with PL/VAE over 0/3 (none), 1/3, 2/3, and all trans-data.  }
    \vspace{-0.5em}
    \label{fig:ana-scale}
     \vspace{-1em}
\end{figure}

%% file: figures/ana-label.tex
\begin{figure}[t]
    \centering
    \includegraphics[width=.4\textwidth]{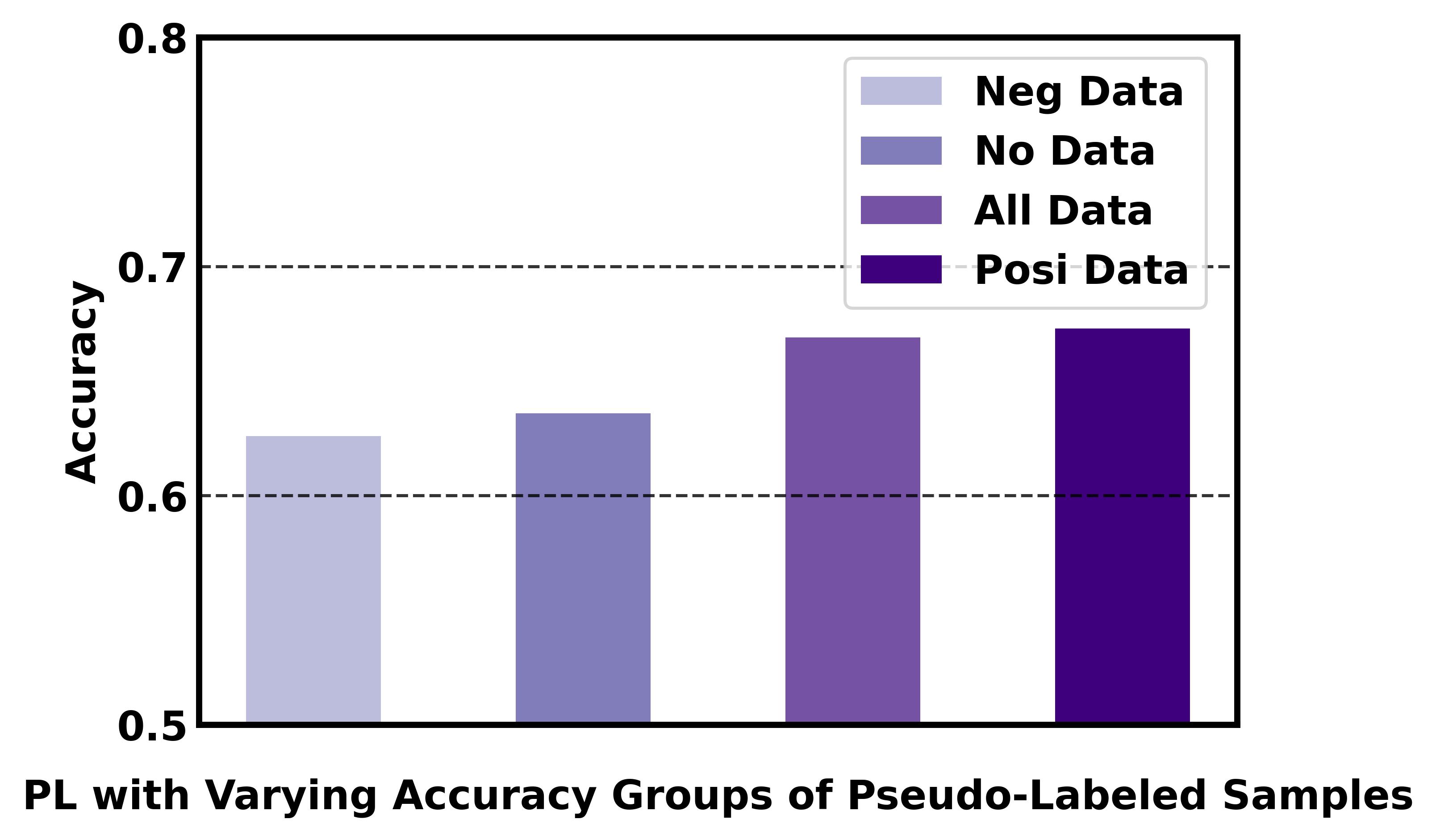}
    \vspace{-0.5em}
    \caption{PL accuracy tested on $t_{4}$ (y-axis). 
    The bars from left to right show PL learned with all negative samples, no sample, all samples, and all positive samples. 
    }
    \vspace{-0.5em}
    \label{fig:ana-label}
    \vspace{-0.5em}
\end{figure}

%% file: figures/ana-time.tex
\begin{figure}[htb]
    \centering
    \includegraphics[width=.35\textwidth]{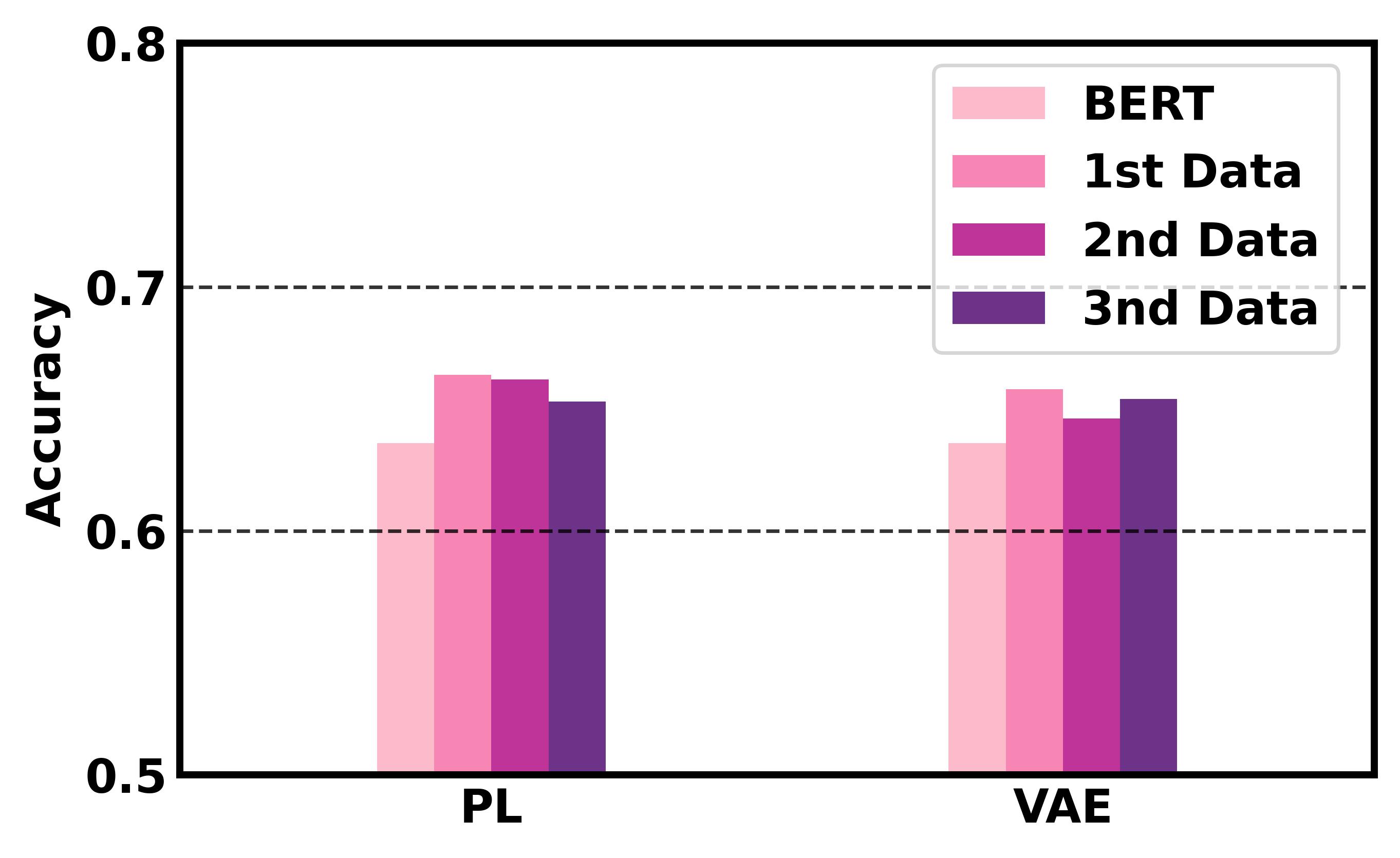}
    \vspace{-0.5em}
    \caption{PL (left group) and VAE (right group) test accuracy on $t_{4}$ (y-axis).
    Within a group, each bar from left to right shows the learning with no trans-data (BERT only) and that from $t_1$, $t_2$, and $t_3$, varying in freshness.
    }
    \vspace{-0.5em}
    \label{fig:ana-time}
    \vspace{-0.5em}
\end{figure}

%% file: figures/heatmap.tex
\begin{figure}[t]
\centering    \includegraphics[width=0.49\textwidth]{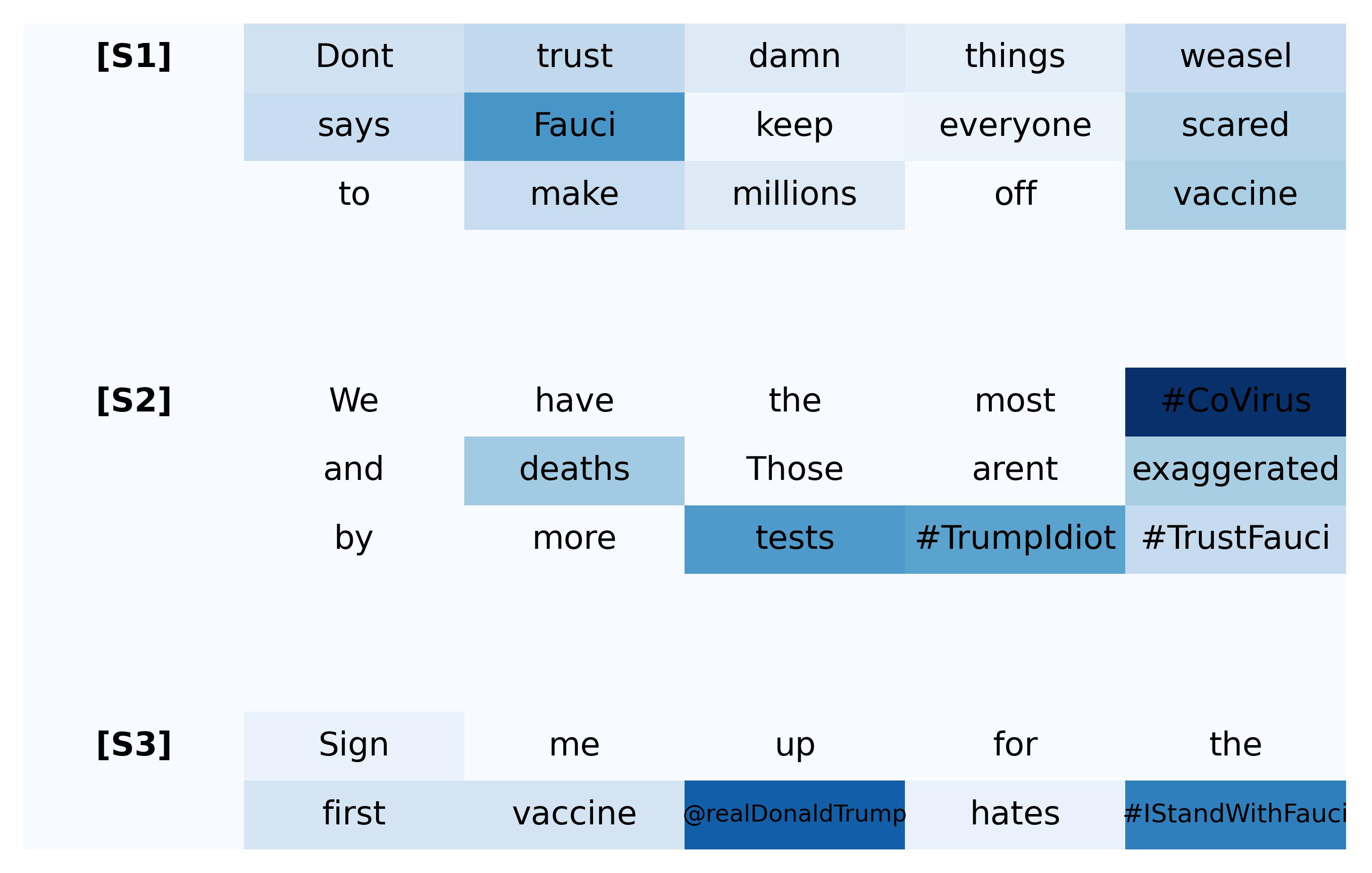}
\vspace{-1em}
\caption{\label{fig:heatmap}
The heatmap visualization of self-attention weights for trans-data samples from $t_1$ ($S_1$), $t_2$ ($S_2$), and $t_3$ ($S_3$). Darker colors indicate higher weights. 
}
\vspace{-1.0em}

\end{figure}

%% file: sections/conclusion.tex
\section{Conclusion}
We have presented an empirical study to substantially experiment NLU in a dynamic social media environment.
Results on four popular tasks in social media classification shed light on the following findings. 
(1) Time evolution indeed exhibits changing features, which may negatively and universally affect social media NLU effectiveness.
(2) Popular UDA baselines PL and VAE are potentially helpful with some trans-data hinting the past-to-future correlations.
(3) Collaborating VAE and PL, varying in pros and cons, exhibits promising preliminary result, although it requires careful research design.
